\documentclass[11pt,a4paper]{article}
\pdfoutput=1
\usepackage[hyperref]{acl2019}
\usepackage{times}
\usepackage{latexsym}
\usepackage{url}
\usepackage{multirow}
\usepackage{graphicx}
\usepackage{float}

\aclfinalcopy 

\title{Integrating Lexical Knowledge in Word Embeddings using Sprinkling and Retrofitting}

\author{Aakash Srinivasan$^{1}$\thanks{$^{*}$Equal Contribution}, Harshavardhan Kamarthi$^{2*}$, Devi Ganesan$^{2}$, Sutanu Chakraborti$^{2}$\\
$^{1}$Dept. of Computer Science, University of California, Los Angeles \\
$^{2}$Dept. of Computer Science and Engineering, Indian Institute of Technology Madras\\
Email: \{s.aakash3431, harshavardhan864.hk\}@gmail.com,
\{gdevi, sutanuc\}@cse.iitm.ac.in,}

\date{}

\begin{document}
\maketitle
\begin{abstract}
Neural network based word embeddings, such as Word2Vec and GloVe, are purely data driven in that they capture the distributional information about words from the training corpus. 
Past works have attempted to improve these embeddings by incorporating semantic knowledge from lexical resources like WordNet. Some techniques like retrofitting modify word embeddings in the post-processing stage while some others use a joint learning approach by modifying the objective function of neural networks. 
In this paper, we discuss two novel approaches for incorporating semantic knowledge into word embeddings. In the first approach, we take advantage of Levy et al's work which showed that using SVD based methods on co-occurrence matrix provide similar performance to neural network based embeddings. We propose a ‘sprinkling’ technique to add semantic relations to the co-occurrence matrix directly before factorization. In the second approach, WordNet similarity scores are used to improve the retrofitting method. We evaluate the proposed methods in both intrinsic and extrinsic tasks and observe significant improvements over the baselines in many of the datasets.
\end{abstract}

\section{Introduction}

Neural Network based models \cite{mikolov2013efficient,GloVe} have been hugely successful in generating useful vector representation for words which preserve their distributional properties in a given corpora.
Improving the quality of word embeddings have led to better performance in many downstream language tasks.
Considering the widespread uses of word embeddings, there have been a lot of interest in improving the quality of these embeddings by leveraging lexical knowledge such as synonymy, hypernymy, hyponymy, troponymy and paraphrase relations. This is accompanied by the availability of large scale lexical knowledge available in WordNet \cite{miller1995wordnet} and Paraphrase Database (PPDB) \cite{ganitkevitch2013ppdb}. 

In this paper, we propose two simple yet powerful approaches to incorporate lexical knowledge into the word embeddings. First, we propose a matrix factorization based approach which uses the idea of `sprinkling' \cite{chakraborti2006sprinkling,chakraborti2007supervised} semantic knowledge into the word co-occurrence matrix. Second, we identify the weaknesses of the retrofitting model \cite{faruqui2014retrofitting} and propose a few modifications that improves the performance. We demonstrate the strength of the proposed models by showing significant improvements in two commonly used intrinsic language tasks - word similarity and analogy, and two extrinsic tasks - named entity recognition (NER) and part of speech tagging (POS).

\section{Related Works}

Learning of word embeddings that capture distributional information has been vital to many NLP tasks. Prediction-based methods such as skip-gram \cite{mikolov2013efficient} and CBOW \cite{bengio2003neural} use neural language modelling for predicting a given word given its context words (or vice-versa) and extract the learned weight vectors as word embeddings. On the other hand, count-based methods derive a co-occurrence matrix of words in the corpus and use matrix factorization techniques like SVD to extract word representations \cite{levy2014neural}. GloVe \cite{GloVe} uses co-occurrence matrix to train word embeddings such that the dot product between any two words is proportional to the log probability of their co-occurrence.

The models that incorporate lexical knowledge into the word embeddings can be broadly classified into two categories, namely post processing and joint learning. Post processing methods such as \cite{faruqui2014retrofitting,mrkvsic2016counter} take the pre-trained word embeddings and modify them by injecting semantic knowledge. 
The retrofitting method \cite{faruqui2014retrofitting} derives similarity constraints from WordNet and other resources to pull similar words closer together. Whereas, the counterfitting approach,  \cite{mrkvsic2016counter} also tries to push the antonymous words away from each other. These approaches consider only one-hop neighbours' relations. We improve upon this by considering multi-hop neighbours as well as use structural and information-based similarity scores to determine their relative importance in imposing similarty contraints to the word embeddings.

Joint learning approaches like \cite{yu2014improving,fried2014incorporating, Vashishth2018IncorporatingSA} learn word embeddings by jointly optimizing distributional and relational information. For instance, in Yu and Dredze \shortcite{yu2014improving}, the objective function consists of  both the original skip-gram objective as well as prior knowledge  from  semantic  resources  to learn  improved  lexical  semantic  embeddings. The recent work by Vashishth et al. \shortcite{Vashishth2018IncorporatingSA} uses Graph Convolutional Networks (GCNs) to learn relations between words and out-performs the previous methods in many language tasks.

\paragraph{Sprinkling:} Latent Semantic Indexing (LSI), also known as Latent Semantic Analysis (LSA), learns a distributional representation for words by performing Singular Value Decomposition (SVD) on the term-document matrix. However, the dimensions obtained from LSI are not optimal in a classification setting because it is agnostic to class label information of the training data. The sprinkling method introduced by Chakraborti et al., \shortcite{chakraborti2006sprinkling} improves LSI by appending the class labels as extra features (terms) to the corresponding training documents. When LSI is carried out on this augmented term-document matrix, terms pertaining to the same class are pulled closer to each other. An extension of this method, called adaptive sprinkling \cite{chakraborti2007supervised}, allows to control the importance of specific class labels by appending them multiple times to the term-document matrix. For instance, in case of double sprinkling, we append the class labels twice to the matrix thus improving the weakly supervised constraints imposed by class labels.

\section{Proposed Models}
In this section, we discuss the proposed models to incorporate semantic knowledge into word embeddings. 

\subsection{SS-PPMI \& DSS-PPMI} 
In this approach, we take advantage of 
Levy and Goldberg's work \shortcite{levy2014neural} in which the authors have shown that the objective function used in Word2vec \cite{mikolov2013efficient} implicitly factorizes a Shifted PPMI (SPPMI) matrix. While there are many methods that attempt to inject semantic knowledge into neural word embeddings, to the best of our knowledge, we have not come across any work that tries to inject semantic knowledge into the SPPMI matrix. In its original form, the SPPMI matrix captures only distributional information. Hence, we are interested in analysing the impact of injecting semantic knowledge into the SPPMI matrix and the effectiveness of the resulting word embeddings.

Inspired from \cite{chakraborti2006sprinkling,chakraborti2007supervised}, which exploits the class knowledge of the documents by 'sprinkling' label terms into the term-document matrix before matrix factorization, we modify the SPPMI matrix by adding reachability information from lexical knowledge bases such as WordNet and PPDB. In the lexical graphs obtained from these knowledge bases, words are connected by edges representing relations such as synonymy, hypernymy, etc. We say that a word $v$ is reachable from another word $u$ if and only if there exists a path between them in the lexical graph. More formally, let $n$ be the size of the vocabulary. We define the reachability matrix $L_k$ $\in$ $\{0,1\}^{n\times n}$ to be a zero-one square matrix with each element $L_k(u,v)$ indicating if word $v$ is reachable  from word $u$ within $k$ hops in the lexical knowledge graph. 

We concatenate the reachability matrix with the SPPMI matrix to obtain \textit{Sprinkled Shifted - Positive PMI} (\textbf{SS-PPMI}). We then perform SVD on this augmented matrix to obtain the enriched word embeddings.
\begin{equation}
    \textit{SPPMI} = max(PMI - \log (neg), 0 )
\end{equation}
\begin{equation}
\textit{SS-PPMI} = \textit{SPPMI} \circ L_k
\end{equation}
\begin{equation}
    \textit{SS-PPMI} \approx U_x \Sigma_x V_{x}^T
\end{equation}
\begin{equation}
    \textit{Embeddings} = U_x \Sigma_x^{p}
\end{equation}
where $\circ$ denotes the matrix concatenation operation, $neg$ denotes the number of negative samples and $x$ denotes the lower rank approximation of the \textit{SS-PPMI} matrix. \textit{SS-PPMI} matrix is of dimensions $n \times 2n$. Following the work of Levy et al., \shortcite{levy2014neural}, we have used $p$ as 0.5 to obtain the word embeddings. 

The original motivation for sprinkling technique \cite{chakraborti2006sprinkling} was that documents of same class are brought closer by appending the class labels to term-document matrix.
Likewise, words which have strong syntactic relations such as synonymy or antonymy have similar neighbourhood in graphs like WordNet. This translates to these word pairs having similar columns in the reachability matrix. Thus, appending reachability matrix to SPPMI matrix would bring such words closer. 

We can further strengthen these constraint by adding the reachability matrix multiple times as done in adaptive sprinkling \cite{chakraborti2007supervised}.
 We performed experiments adding reachability matrix twice and we call the resulting matrix as \textit{Doubly Sprinkled Shifted - Positive PMI} (\textbf{DSS-PPMI}), which will be of dimensions $n\times 3n$.

\subsection{W-Retrofitting}
Retrofitting was introduced by Faruqui et al., \shortcite{faruqui2014retrofitting} and is a method to add semantic information to pre-trained word vectors. The post-processing step modifies the word embeddings such that the embeddings of words with semantic relations between them are pulled towards each other. Formally, given the pre-trained vectors $\hat{Q}$ = $(\hat{q_1},\hat{q_2}\cdots \hat{q_n})$, and a knowledge base represented by the adjacency matrix $A$, we need to learn new vectors $Q$ = $(q_1,q_2\cdots q_n)$ such that following objective $\psi(Q)$ is minimized:
\begin{equation}
    \psi(Q) = \sum_{i=1}^{i=n} (\alpha_i \| q_i - \hat{q_i} \|^2 + \sum_{j=1}^{j=n}\beta_{ij}A_{ij} \| q_i - q_j \|^2)
\label{eqn:retro}
\end{equation}
The objective is a convex function and we can find the solution using the efficient iterative update method used in Faruqui et al., \shortcite{faruqui2014retrofitting}:
\begin{equation}
    q_i = \frac{\sum_{j=1}^{j=n}A_{ij}\beta_{ij}q_j + \alpha_i q_i}{\sum_{j=1}^{j=n}A_{ij}\beta_{ij} + \alpha_i}
\end{equation}

 The $\beta_{ij}$ term is usually assigned as $degree(i)^{-1}$. This choice of assigning weights can be done in a better way by learning from semantic knowledge sourcea such as WordNet. 
 
 We propose a modification to the retrofitting methods called \textbf{W-Retrofitting} (weighted retrofitting), where we use WordNet-based similarity scores to obtain a better setting of $\beta_{ij}$. For two words $w_i$ and $w_j$ with WordNet similarity score $Sim(i,j)$, $\beta_{ij}$ is obtained by normalizing the similarity scores across neighbors  and is given as: $\beta_{ij}=\frac{Sim(i,j)}{\sum_{j'}Sim(i,j')}$. Since a word can have multiple synsets, the similarity score is the maximum of the similarity scores of all possible pairs of synsets, taking one each from the two words. For information based similarity measures like Lin similarity we compute mutual information from a random subset of Wikipedia corpus containing 100,000 articles. Further, we extend our method to consider nodes which are atmost 2 hops from given node when computing weights.
 \begin{table}[]
\centering
\begin{tabular}{|c|c|}
\hline
Scores                           & Datasets                 \\ \hline
Similarity                       & RG65, WS353S, Simlex-999 \\ 
Relatedness                      & WS353R, TR9856           \\ 
No Distinguishing & MEN, RW, MTunk, WS353    \\ \hline
\end{tabular}
\caption{The characterization of scores given by different word similarity datasets}
\label{tab:word_similarity_datasets}
\end{table}

\section{Experimental Setup}

\subsection{Intrinsic Evaluation}
We evaluate the proposed models on word similarity and analogy tasks.

\noindent\textbf{Word similarity}: We use MEN \cite{bruni2014multimodal}, MTunk \cite{radinsky2011word}, RG65 \cite{rubenstein1965contextual}, Rare Words(RW) \cite{luong2013better}, SimLex999 \cite{hill2015simlex}, TR9856 \cite{levy2015tr9856}, WS353 \cite{finkelstein2002placing}, WS353S (Similarity), WS353R (Relatedness). Spearman correlation is used as evaluation metric.

\noindent\textbf{Analogy}: We evaluated analogy task with Google Analogy \cite{mikolov2013efficient}, MSR Analogy \cite{mikolov2013linguistic} and Semeval2012 datasets. We follow the standardized setup as explained in \cite{jastrzebski2017evaluate}. 

\subsection{Sources of Knowledge}

We used two sources of semantic knowledge: WordNet \cite{miller1995wordnet} and PPDB \cite{ganitkevitch2013ppdb}. We used the same PPDB knowledge source used in Faruqui et al., \shortcite{faruqui2014retrofitting}. We used WordNet source knowledge from V. Batagelj \shortcite{pajekwn}. The relations considered are synonymy, hypernymy, meronymy and verb entailment. PPDB has 84467 nodes and 169703 edges, WordNet source we used has 82313 nodes and 98678 edges.

We used the latest Wikipedia dump\footnote{https://dumps.wikimedia.org/enwiki/latest/} containing 6 Billion wikipedia articles to generate the SPPMI matrix. We followed the same procedure as given in Levy et al., \shortcite{levy2015improving} and chose the number of negative samples to be default value of 5. In all of our experiments, we chose embedding dimension as 300, which is commonly used in the literature.

\subsection{Baselines}
We use the following baselines for comparison
\begin{enumerate}
    \item \textbf{GloVe}: Our first baseline is the GloVe embeddings \cite{GloVe} trained on the Wikipedia corpus retrieved from Stanford NLP group website\footnote{https://nlp.stanford.edu/projects/GloVe/}.
    \item \textbf{Retrofit}: We apply the retrofitting technique \cite{faruqui2014retrofitting} on the GloVe embeddings where Wordnet or PPDB was as the source of word relations.
    \item \textbf{SPPMI}: We perform SVD on the Shifted PPMI matrix (as mentioned in Section 3) without sprinkling.
    
    \item \textbf{SynGCN} \cite{Vashishth2018IncorporatingSA}: This work uses Graph-convolution based methods to impart relational information between words and have shown state-of-art results in many benchmarks. We directly report the available results from the original paper which uses same evaluation benchmarks. 
\end{enumerate}

\subsection{Extrinsic Evaluation}
\label{sec:extr}
To further test the effectiveness of the different methods in grounding word meanings, we utilize the embeddings in following tasks. The neural network architectures used for each of the tasks are same as that used in Vashishth et al., \shortcite{Vashishth2018IncorporatingSA}.

\begin{enumerate}
\item \textbf{Part-of-speech tagging (POS):} This task classifies each word of given sentence as one of the part-of-speech tags. We use the LSTM based neural architecture discussed in Reimers and Gurevych \shortcite{Reimers2017ReportingSD} on the Penn treebank dataset  \cite{Marcus1994ThePT}.

\item \textbf{Named-entity recognition (NER):} The goal of this task is to extract and classify named entities in the sentences as person, organisation, location or miscellaneous. We use the model proposed in Lee et al., \shortcite{Lee2018HigherorderCR} on CoNLL-2003 dataset \cite{Sang2003IntroductionTT}.
\end{enumerate}

\section{Results and Analysis}

\subsection{SS-PPMI}
\textbf{Reachability Matrix is powerful in capturing semantic information}: We proposed a simple sprinkling approach in which a zero-one matrix captures the $k$-hop reachability information between words in a lexical knowledge graph. In order to see how effectively the reachability matrix captures the lexical knowledge, we performed SVD on the reachability matrix and obtained the word embeddings. Table \ref{tab:label_matrix_wordsim} shows the performance of the obtained embeddings on word similarity task, The dimension of embedding used is 300. Interestingly, we clearly observe that the embeddings obtained from the reachability matrix only (without SPPMI matrix) compete strongly with 300 dimensional pretrained GloVe embeddings on the similarity based datasets. The best performing model gives  a Spearman correlation which is \textbf{0.19} more than GloVe in Simlex999. Similarly, in RG65 and WS353S, the reachability based embeddings compete well with GloVe. Between the choice of PPDB or WordNet as the lexical knowledge sources, PPDB seems to be more helpful. In general, the performance of reachability-based embeddings increases with increasing the number of hops on the similarity datasets.

In the case of relatedness datasets, the model competes poorly with the baseline-GloVe. This is quite expected as the reachability matrix doesn't capture any information about the word co-occurrence. These observations have been foundational to our proposed \textit{SS-PPMI} and \textit{DSS-PPMI} methods.

\begin{table*}[t!]
\centering
\scalebox{0.8}{
\begin{tabular}{|c|c|c|c|c|c|c|c|c|c|c|}
\hline
                         &          & \multicolumn{3}{c|}{Similarity}                  & \multicolumn{2}{c|}{Relatedness} & \multicolumn{4}{c|}{No Distinction}                               \\ \hline
Lexical Knowledge        & Hops - k & SimLex999      & WS353S         & RG65           & WS353R          & TR9856         & WS353          & MEN            & MTurk          & RW             \\ \hline
Baseline - GloVe         & -        & 0.370          & \textbf{0.665} & \textbf{0.769} & \textbf{0.560}  & \textbf{0.575} & \textbf{0.601} & \textbf{0.737} & \textbf{0.633} & \textbf{0.411} \\ \hline
\multirow{2}{*}{PPDB}    & 1        & 0.507          & 0.461          & 0.433          & 0.127           & 0.273          & 0.336          & 0.284          & 0.181          & 0.465          \\ \cline{2-11} 
                         & 2        & \textbf{0.529}          & 0.567          & 0.512          & 0.128           & 0.261          & 0.362          & 0.304          & 0.261          & 0.506          \\ \hline 
\multirow{2}{*}{WordNet} & 1        & 0.077          & 0.343          & 0.110          & 0.151           & 0.128          & 0.293          & 0.161          & 0.063          & 0.056          \\ \cline{2-11} 
                         & 2        & 0.209          & 0.349          & 0.378          & 0.163           & 0.149          & 0.285          & 0.275          & 0.145          & 0.209          \\ \hline
\end{tabular}
}
\caption{Performance of the reachability-based embeddings on similarity datasets. Reported numbers are the Spearman correlation coefficients.}
\label{tab:label_matrix_wordsim}
\end{table*}

\begin{table*}[t]
\centering
\scalebox{0.7}{
\begin{tabular}{|c|c|c|c|c|c|c|c|c|c|c|c|}
\hline
                          &                          &      & \multicolumn{3}{c|}{Similarity}                  & \multicolumn{2}{c|}{Relatedness} & \multicolumn{4}{c|}{No Distinction}                               \\ \hline
Method                    & Lexical Knowledge        & hops & SimLex999      & WS353S         & RG65           & WS353R          & TR9856         & WS353          & MEN            & MTurk          & RW             \\ \hline
SPPMI                     & -                        & -    & 0.385          & 0.728          & 0.783          & 0.603           & 0.625          & 0.663          & 0.742          & 0.599          & 0.516          \\ \hline
SynGCN & - & -                                                      & 0.455           & \textbf{0.732}         & -             & 0.457              & -             & 0.601     & -     & -         & 0.337 \\ \hline
\multirow{2}{*}{SS-PPMI}  & \multirow{2}{*}{PPDB}    & 1    & 0.386          & 0.728          & 0.782          & 0.604           & 0.625          & 0.663          & 0.742          & 0.599          & 0.516          \\ \cline{3-12} 
                          &                          & 2    & 0.398          & 0.733          & 0.775          & 0.619           & 0.628          & 0.669          & 0.743          & 0.610          & \textbf{0.521}          \\ \hline
\multirow{2}{*}{DSS-PPMI} & \multirow{2}{*}{PPDB}    & 1    & 0.386          & 0.728          & 0.782          & 0.604           & 0.625          & 0.663          & 0.742          & 0.599          & 0.516          \\ \cline{3-12} 
                          &                          & 2    & \textbf{0.420}          & 0.733          & 0.780          & 0.620           & \textbf{0.629} & 0.668          & 0.743          & 0.607          & 0.528          \\ \hline 
\multirow{2}{*}{SS-PPMI}  & \multirow{2}{*}{WordNet} & 1    & 0.393          & 0.724          & 0.792          & 0.627           & 0.597          & 0.667          & 0.769          & 0.611          & 0.464          \\ \cline{3-12} 
                          &                          & 2    & 0.394          & 0.733          & 0.793          & 0.629           & 0.601          & 0.671          & 0.770          & 0.616          & 0.435          \\ \hline 
\multirow{2}{*}{DSS-PPMI} & \multirow{2}{*}{WordNet} & 1    & 0.393          & 0.724          & 0.792          & 0.627           & 0.597          & 0.667          & 0.769          & 0.611          & 0.463          \\ \cline{3-12} 
                          &                          & 2    & 0.394          & \textbf{0.739}          & \textbf{0.804}          & \textbf{0.638}  & 0.599          & \textbf{0.677}          & \textbf{0.771} & \textbf{0.619} & 0.414          \\ \hline
\end{tabular}
}
\caption{Results on word similarity datasets using SS-PPMI and DSS-PPMI embeddings}
\label{tab:wordsim_ssppmi}
\end{table*}

\noindent\textbf{SS-PPMI and DSS-PPMI provide significant improvements in word similarity and analogy:} Table \ref{tab:wordsim_ssppmi} provides the results with \textit{SS-PPMI} and \textit{DSS-PPMI} approaches on word similarity task with embedding dimension as 300. We clearly observe that the proposed models defeat the baseline in all the datasets. The margin of improvement is quite high in case of similarity datasets. We see close to \textbf{0.21} increase in spearman correlation for Simlex999, \textbf{0.04} increase in RG65. This is somewhat expected as we already saw that reachability matrix contains lexical information. Interestingly, we also saw improvements in relatedness datasets where the sprinkling approaches perform narrowly better than SPPMI based approach. In other datasets like WS353, MEN we see improvements of about 0.02 and 0.03 in spearman correlation respectively. Overall, sprinkling significantly improves the performance on word similarity task.

Overall, we observe that Double Sprinkling method (\textit{DSS-PPMI}) works better than \textit{SPPMI} in word similarity task. Increasing the number of hops ($k$) in the reachability matrix improves the performance in word similarity , in general.

Table \ref{tab:analogy_ssppmi} shows improvements provided by the sprinkling methods on analogy datasets. We observe marginal improvements over baseline in google and SemEval2012. 

\begin{table}[h]
\centering
\scalebox{0.7}{
\begin{tabular}{|c|c|c|c|c|}
\hline
Method                    & Graph        & hops & Google                     & SemEval \\ \hline
SPPMI-Baseline            & -                        & -    & 0.337                   & 0.176          \\ \hline
SynGCN          &                       &       &   -   &   \textbf{0.234}
    \\ \hline
\multirow{2}{*}{SS-PPMI}  & \multirow{2}{*}{PPDB}    & 1    & 0.338                    & 0.175          \\ \cline{3-5} 
                          &                          & 2    & \textbf{0.347}           & \textbf{0.180} \\ \hline
\multirow{2}{*}{DSS-PPMI} & \multirow{2}{*}{PPDB}    & 1    & 0.338                   & 0.176          \\ \cline{3-5} 
                          &                          & 2    & 0.343                    & 0.188          \\ \hline
\multirow{2}{*}{SS-PPMI}  & \multirow{2}{*}{WordNet} & 1    & 0.122                   & 0.166          \\ \cline{3-5} 
                          &                          & 2    & 0.121                    & 0.165          \\ \hline 
\multirow{2}{*}{DSS-PPMI} & \multirow{2}{*}{WordNet} & 1    & 0.122                    & 0.166          \\ \cline{3-5} 
                          &                          & 2    & 0.118                    & 0.161          \\ \hline 
\end{tabular}
}
\caption{Analogy results using proposed SS-PPMI and DSS-PPMI approaches}
\label{tab:analogy_ssppmi}
\end{table}

\subsection{W-Retrofitting}
We apply our W-retrofitting model to GloVe \cite{GloVe} embeddings trained on Wikipedia corpus. We experimented with one hop and two hop neighbors and several methods for similarity estimation: inverse path similarity, Jaing-Conrath Similarity \cite{jiang1997semantic}, Wu -Palmer Similarity \cite{wu1994verbs}, Leacock-Chowdorov Similarity \cite{leacock1998combining} and Lin Similarity \cite{lin1998information}. The neighbourhood information for estimating similarity was obtained from either WordNet or PPDB graphs. \textit{We found that Jaing-Conrath Similarity works best for WordNet,  inverse path similarity for PPDB}. So, we report results for these similarity measures only.

\paragraph{Word Similarity:} The performances of all our models are either comparable or superior to baselines as seen in table \ref{tab:retro_ws}. We see that using PPDB knowledge source and path based similarity as weights in the retrofit objective functions gives the best performance and outperforms the baselines in most benchmarks.

\begin{table*}[!t]
\centering
\scalebox{0.68}{
\begin{tabular}{|l|l|l|l|l|l|l|l|l|l|l|l|}
\hline
                                  &                          & \multicolumn{4}{c|}{Similarity}                        & \multicolumn{3}{c|}{Relatedness}                 & \multicolumn{3}{c|}{No Distinction}              \\ \hline
Method                            & Lexical Knowledge        & Hops & SimLex999      & WS353S        & RG65           & WS353R         & TR9856         & MTurk          & WS353          & MEN            & RW             \\ \hline
GloVe-baseline                    &                          & -    & 0.37           & 0.665         & 0.769          & 0.56           & 0.575          & 0.633          & 0.601          & 0.737          & 0.411          \\ \hline
SynGCN & - & -                                                      & 0.455           & \textbf{0.732}         & -             & 0.457              & -             & 0.601     & -     & -         & 0.337 \\ \hline

Retrofit-baseline                 & \multirow{3}{*}{PPDB}    & 1    & 0.496          & 0.7           & \textbf{0.825} & \textbf{0.585} & \textbf{0.601} & \textbf{0.675} & 0.631          & 0.764          & \textbf{0.431} \\ \cline{1-1} \cline{3-12} 
\multirow{2}{*}{W-retrofit(path)} &                          & 1    & \textbf{0.509} & \textbf{0.71} & 0.824          & 0.583          & 0.584          & 0.669          & \textbf{0.641} & \textbf{0.773} & 0.417          \\ \cline{3-12} 
                                  &                          & 2    & 0.422          & 0.628         & 0.788          & 0.519          & 0.525          & 0.63           & 0.562          & 0.722          & 0.372          \\ \hline
Retrofit-baseline                 & \multirow{3}{*}{Wordnet} & 1    & \textbf{0.434} & 0.693         & 0.774          & \textbf{0.557} & 0.574          & \textbf{0.642} & 0.607          & \textbf{0.766} & 0.387          \\ \cline{1-1} \cline{3-12} 
\multirow{2}{*}{W-retrofit(jcn)} &                          & 1    & 0.432          & 0.685         & 0.772          & 0.543          & 0.568          & 0.64           & 0.6            & 0.764          & 0.353          \\ \cline{3-12} 
                                  &                          & 2    & 0.399          & \textbf{0.73} & \textbf{0.785} & 0.528          & \textbf{0.579} & 0.634          & \textbf{0.616} & 0.764          & \textbf{0.389} \\ \hline
\end{tabular}
}
\caption{Word Similarity results for W-Retrofitting approach}
\label{tab:retro_ws}
\end{table*}

\paragraph{Analogy:} Some of our models outperform retrofitting baselines in Google analogy. In SemEval task, we mostly outperform GloVe but retrofitting baseline on WordNet gives the best score. The results are summarised in table \ref{tab:retro_analogy}

\begin{table}[h]
\centering
\scalebox{0.8}{
\begin{tabular}{|l|l|l|l|l|}
\hline
Similarity        & Graph & Hops & Google         & SemEval \\ \hline
GloVe             &                   & 0    & \textbf{0.717} & 0.164          \\ \hline
SynGCN          &                       &       &   -   &   \textbf{0.234}
    \\ \hline
Retrofit-baseline & \multirow{3}{*}{PPDB}              & 1    & 0.451          & 0.171          \\ \cline{3-5} \cline{1-1}
\multirow{2}{*}{path}                 &               & 1    & 0.448          & 0.167          \\ \cline{3-5}
                  &                   & 2    & 0.248          & 0.151          \\ \hline
Retrofit-baseline & \multirow{3}{*}{WordNet}           & 1    & 0.603          & \textbf{0.184} \\ \cline{3-5} \cline{1-1}

\multirow{2}{*}{jcn}               &            & 1    & \textbf{0.701} & 0.161          \\ \cline{3-5}
                  &            & 2    & 0.693          & 0.155          \\ \hline
\end{tabular}
}
\caption{Analogy results for W-Retrofitting}
\label{tab:retro_analogy}
\end{table}

\begin{table}[h]
    \centering
    \scalebox{0.8}{
    \begin{tabular}{|c|c|c|c|}
    \hline
    Model               & SimLex999      & WS353S         & RG65           \\ \hline
    SPPMI               & 0.276          & 0.624          & 0.671          \\ \hline
    Retrofitting        & 0.336          & 0.624          & 0.752          \\ \hline
    W-Retrofitting      & 0.429          & 0.656          & 0.747          \\ \hline
    Reachability Matrix & 0.561          & 0.567          & 0.664          \\ \hline
    Sprinkling          & \textbf{0.591} & \textbf{0.748} & \textbf{0.821} \\ \hline\hline
    Model               & WS353R         & TR9856         & MTurk          \\ \hline
    SPPMI               & 0.509          & 0.527          & 0.626          \\ \hline
    Retrofitting        & 0.479          & 0.534          & 0.623          \\ \hline
    W-Retrofitting      & 0.521          & 0.548          & \textbf{0.631} \\ \hline
    Reachability Matrix & 0.194          & 0.325          & 0.283          \\ \hline
    Sprinkling          & \textbf{0.638} & \textbf{0.629} & 0.619          \\ \hline\hline
    Model               & WS353          & MEN            & RW             \\ \hline
    SPPMI               & 0.562          & 0.691          & 0.359          \\ \hline
    Retrofitting        & 0.545          & 0.708          & 0.350          \\ \hline
    W-Retrofitting      & 0.595          & 0.726          & 0.384          \\ \hline
    Reachability Matrix & 0.376          & 0.325          & 0.506          \\ \hline
    Sprinkling          & \textbf{0.682} & \textbf{0.771} & \textbf{0.560} \\ \hline
        
    \end{tabular}
    }
    \caption{Comparison with various baselines for word similarity and relatedness.}
    \label{tab:overall_analysis}
\end{table}

\begin{table}[h]
\centering
    \scalebox{0.8}{
\begin{tabular}{|l|l|l|l|l|}
\hline
Method                    & Graph                    & Hops & NER           & POS           \\ \hline
SPPMI-Baseline            &                          &      & 82.3          & 92.9          \\ \hline
\multirow{2}{*}{SS-PPMI}  & \multirow{2}{*}{PPDB}    & 1    & 83.4          & 93.3          \\ \cline{3-5} 
                          &                          & 2    & \textbf{84.7} & 93.4          \\ \hline
\multirow{2}{*}{DSS-PPMI} & \multirow{2}{*}{PPDB}    & 1    & 82.3          & \textbf{93.5} \\ \cline{3-5} 
                          &                          & 2    & 87.3          & 93.4          \\ \hline
\multirow{2}{*}{SS-PPMI}  & \multirow{2}{*}{Wordnet} & 1    & 83.5          & 92.8          \\ \cline{3-5} 
                          &                          & 2    & 83.9          & 93.2          \\ \hline
\multirow{2}{*}{DSS-PPMI} & \multirow{2}{*}{Wordnet} & 1    & 83.2          & 93.2          \\ \cline{3-5} 
                          &                          & 2    & 83.5          & 93.1          \\ \hline
\end{tabular}
}
\caption{Results on Extrinsic Evaluation tasks using SS-PPMI and DSS-PPMI embeddings}
\label{tab:extrinsic_spr}
\end{table}

\begin{table}[h]
\centering
    \scalebox{0.8}{
\begin{tabular}{|l|l|l|l|l|}
\hline
Method                & Graph                    & Hops & NER           & POS           \\ \hline
GloVe                 & -                        &      & 89.1          & 94.6          \\ \hline
SynGCN                & -                        &      & \textbf{89.5} & \textbf{95.4} \\ \hline
Retrofit-baseline     & \multirow{3}{*}{PPDB}    & 1    & 88.8          & 94.8          \\ \cline{1-1} \cline{3-5} 
\multirow{2}{*}{path} &                          & 1    & 88.7          & 95            \\ \cline{3-5} 
                      &                          & 2    & 89.2          & 95.1          \\ \hline
Retrofit-baseline     & \multirow{3}{*}{Wordnet} & 1    & 88.2          & 94.5          \\ \cline{1-1} \cline{3-5} 
\multirow{2}{*}{jcn}  &                          & 1    & 88.9          & 95            \\ \cline{3-5} 
                      &                          & 2    & \textbf{89.4} & \textbf{95.3} \\ \hline
\end{tabular}
}

\caption{Results on Extrinsic Evaluation tasks using W-Retrofitting}
\label{tab:extrinsic_retro}
\end{table}

\subsection{Overall Comparison on Word Similarity}
In order to make fair and direct comparison between Sprinkling and Retrofitting, we applied retrofitting and W-retrofitting (using inverse-path similarity over PPDB graph) on the 300 dimensional SPPMI vectors. Table \ref{tab:overall_analysis} provides the best results of the models on each of the word similarity and analogy datasets. We make the following observations. W-Retrofitting does much better than Retrofitting in similarity datasets, as what we saw with GloVe embeddings. 
The source of the improvement comes comes from two things: inclusion of two-hop neighbor information and the intelligent choice of weights from WordNet in W-Retrofitting. 

Using only the Reachability Matrix provides very good scores in similarity based datasets, but doesn't capture relatedness information at all. Using sprinkling approach, we manage to obtain embeddings that have optimal combination of similarity and relatedness information and this makes it perform better than all the other baselines in similarity, relatedness and analogy tasks. 

\subsection{Evaluation on Extrinsic tasks}
The results on extrinsic tasks (discussed in Section \ref{sec:extr}) are given in Tables \ref{tab:extrinsic_spr} and \ref{tab:extrinsic_retro}.  In the case of sprinkling methods, we see that there is a clear increase in scores for both the extrinsic tasks from using the proposed SS-PPMI matrix over using only the SPPMI matrix. We also see that models using PPDB perform better. One reason why we do not compare scores of sprinkling based methods with that of GloVe and Retrofitting based ones is that the vocabulary size(number of nodes) in PPDB or Wordnet graphs are lower than that for GloVe. We also didn't consider punctuation symbols in SPPMI unlike GloVe.

In the case of W-retrofitting, scores from the proposed W-Retrofitting model using jcn weights on wordnet graph are very similar to SynGCN model inspite of SynGCN being a more complex model with a lot of hyperparameters. We also see that the other methods of W-retrofitting have comparable performance to SynGCN. We observe improved performance by considering upto 2 hop neighbours over methods considering just 1 hop neighbours. It is quite interesting to see that the proposed light-weight retrofitting model competes strongly with the more complex SynGCN method as shown by the results in Table 9.

\section{Conclusion and Future Work}
In this paper, we proposed two simple yet powerful approaches to incorporate lexical knowledge into word embeddings. The first approach is a matrix factorization method that `sprinkles' higher order graph information into the word co-occurrence and we show that it significantly improves the quality of the word embeddings. Second, we proposed a simple modification to the retrofitting method that improves it performance visibly. We showed the improvements of the proposed models over baselines in a variety of word similarity and analogy tasks, and across two popular lexical knowledge bases.

For extrinsic tasks, W-retrofitting showed comparable performance to the state-of-art  SynGCN model, \cite{Vashishth2018IncorporatingSA} inspite of SynGCN being a more sophisticated model with lots of parameters that constitute the weights of Graph Convolutional layers and linear layers of neural network used as well as many hyperparameters needed for training the neural network (such as number of GCN layers and their dimensions, learning rate, number of epochs, etc.).

In our sprinkling approach, we didn't consider any importance weighting for different relations. One promising direction that can be experimented in future is to use wordnet similarity scores or a combination of co-occurrence and lexical information as importance values in the reachability matrix. We could also use `adaptive sprinkling' \cite{chakraborti2007supervised} to give more importance to relations of specific sets of words.

 The more recent methods that achieve the state-of-art results in a variety of language tasks utilize pre-trained models such as Elmo \cite{peters2018deep}, BERT \cite{Devlin2018BERTPO} and XLNet \cite{Yang2019XLNetGA}. These models that learn context dependent word embeddings are pre-trained for different language tasks and are later fine-tuned for specific tasks. Another direction of research we would like to explore is to study the improvements gained by using our proposed models to initialize the word embeddings before pre-training these models.

\bibliography{references}
\bibliographystyle{acl_natbib}
\end{document}